%% file: main.tex
\documentclass[]{spie}  

\usepackage{amsmath,amsfonts,amssymb}
\usepackage{accents}
\newlength{\dhatheight}
\newcommand{\doublehat}[1]{%
    \settoheight{\dhatheight}{\ensuremath{\hat{#1}}}%
    \addtolength{\dhatheight}{-0.35ex}%
    \hat{\vphantom{\rule{1pt}{\dhatheight}}%
    \smash{\hat{#1}}}}
\usepackage{graphicx}
\usepackage{float}
\usepackage[colorlinks=true, allcolors=blue]{hyperref}

\title{Patch distribution modeling framework adaptive cosine estimator (PaDiM-ACE) for anomaly detection and localization in synthetic aperture radar imagery}

\author[a]{Angelina Ibarra}
\author[a]{Joshua Peeples}
\affil[a]{Department of Electrical and Computer Engineering, Texas A\&M University, College Station, TX, 77845}

\authorinfo{Further author information: (Send correspondence to J.P.)\\J.P.: E-mail: jpeeples@tamu.edu, Telephone: 1 979 458 7026}

\begin{document} 
\maketitle

\begin{abstract}
This work presents a new approach to anomaly detection and localization in synthetic aperture radar imagery (SAR), expanding upon the existing patch distribution modeling framework (PaDiM). We introduce the adaptive cosine estimator (ACE) detection statistic. PaDiM uses the Mahalanobis distance at inference, an unbounded metric. ACE instead uses the cosine similarity metric, providing bounded anomaly detection scores. The proposed method is evaluated across multiple SAR datasets, with performance metrics including the area under the receiver operating curve (AUROC) at the image and pixel level, aiming for increased performance in anomaly detection and localization of SAR imagery. The code is publicly available: \url{https://github.com/Advanced-Vision-and-Learning-Lab/PaDiM-ACE}.
\end{abstract}


\input{sections/Introduction}

\input{sections/Related_Work}

\input{sections/Method}

\input{sections/Results_Discussion}

\input{sections/Conclusion}

\acknowledgments 
 
This work was supported by the Laboratory Directed Research and Development program at Sandia National Laboratories, a multimission laboratory managed and operated by National Technology and Engineering Solutions of Sandia LLC, a wholly owned subsidiary of Honeywell International Inc. for the U.S. Department of Energy’s National Nuclear Security Administration under contract DE-NA0003525.

\bibliography{report} 
\bibliographystyle{spiebib} 

\end{document}

%% file: sections/Introduction.tex
\section{Introduction}
\label{sec:intro}  

Synthetic aperture radar (SAR) is a powerful remote sensing technology that creates high-resolution imagery day or night, regardless of weather conditions\cite{10547107}. SAR's ability to penetrate cloud cover and operate independently of illumination makes it particularly valuable in scenarios where traditional imaging methods may fail\cite{liu2022high}. For this reason, SAR imagery is widely utilized in various applications, including environmental monitoring\cite{9852257}, defense and security\cite{10282026}, and disaster response\cite{ALALI2022103295}. Anomaly detection, the identification of objects or regions that deviate from expected background statistics, is a critical task for such applications. While it has been widely researched in industrial and medical domains, there is a lack of research in anomaly detection in SAR imagery\cite{10283391}. One issue that prevails is the lack of publicly available labeled datasets\cite{li2024sardet100kopensourcebenchmarktoolkit}; therefore, unsupervised and self-supervised approaches have become increasingly popular.  

A promising approach to anomaly detection is Patch Distribution Modeling (PaDiM)\cite{defard2020padimpatchdistributionmodeling}, which leverages pre-trained convolutional neural networks (CNNs) to extract patch-level feature embeddings and models their distribution using a multivariate Gaussian assumption. During inference, the Mahalanobis distance is used to measure deviations from normal data distributions. While effective, PaDiM utilizes the Mahalanobis distance at inference, which is an unbounded metric that can lead to outlier sensitivity\cite{Li_Deng_Li_Jiang_2019}. These limitations motivate the need for an improved framework that enhances generalization.

In this work, we introduce the Adaptive Cosine Estimator (ACE) within the PaDiM framework, resulting in our proposed method, PaDiM-ACE. Instead of relying on the Mahalanobis distance, ACE employs a cosine similarity-based metric for anomaly detection. We evaluate PaDiM-ACE across multiple SAR datasets and present the results of various ablation studies. 

%% file: sections/Related_Work.tex
\section{Related Work}
\label{sec:related_work}  

Anomaly detection methods may be broadly categorized into statistical, classic machine learning, and deep learning approaches\cite{10730574}. However, many of the most effective models rely on deep learning approaches. Anomalib \cite{akcay2022anomalib}, a deep learning library, has implemented various state-of-the-art anomaly detection models, including PaDiM \cite{defard2020padimpatchdistributionmodeling}. PaDiM is a patch distribution modeling framework\cite{defard2020padimpatchdistributionmodeling}. The framework utilizes a pre-trained CNN backbone to extract feature embeddings at different semantic levels for each patch across a batch of images and randomly reduces the embeddings to a size of $d$ features. PaDiM then concatenates the embeddings, assumes a multivariate Gaussian distribution, learns the Gaussian parameters (\textit{i.e.}, mean vector and covariance matrix), and generates a parameter matrix corresponding to each patch's spatial location of the normal training images. At inference, the framework uses the Mahalanobis distance to measure the difference between an input image patch embedding and the learned Gaussian distribution. The distance is used as an anomaly score for each patch to create an anomaly score map. Higher scores indicate anomalous areas, and the max anomaly map score is used to score the entire image. While PaDiM performs well on RGB imagery, we seek to analyze its performance on SAR imagery. 

Although research in anomaly detection in SAR imagery has not been as extensively studied as anomaly detection in industrial or medical domains, several popular methods do exist. One of the most common traditional approaches to anomaly detection in SAR imagery is the use of the Reed-Xiaoli detector (RX)\cite{1990ITASS..38.1760R}, which estimates the background distribution and detects anomalies as deviations. More recently, deep learning-based methods, particularly autoencoders and their variants, have gained popularity. Autoencoders learn a compact representation of normal data and flag deviations as anomalies, while adversarial autoencoders (AAEs) enhance robustness by incorporating adversarial training\cite{9987646}. Hybrid approaches have also been proposed that combine autoencoder-based reconstruction with apriori information from the RX detector to improve anomaly detection performance\cite{10283391}. Due to the lack of labeled SAR datasets, self-supervised learning approaches are promising for anomaly detection in SAR imagery due to their ability to learn meaningful representations without requiring labeled data. One particular approach \cite{9987646} proposes a self-supervised framework that addresses key challenges in SAR anomaly detection, such as speckle noise and spatial correlation structures. Their method first applies SAR2SAR despeckling to enhance image quality, followed by an AAE that reconstructs anomaly-free SAR images. Anomalies are then detected through change detection between input and reconstructed images.

%% file: sections/Method.tex
\section{Method}
\label{sec:method}  

ACE is a binary classifier derived from the generalized likelihood ratio test (GLRT). ACE introduces a target class representative, $\mathbf{s}$, and measures the cosine similarity between a sample feature vector, $\mathbf{x}$, and the target class representative in a whitened coordinate space where the background class distribution is parameterized by the mean vector ($\mathbf{\mu}_b$) and inverse covariance matrix ($\mathbf{\Sigma}_b^{-1}$) as shown in Equation \ref{eqn:ACE}:
\begin{equation}
    D_{ace}=\frac{\mathbf{s}^T\mathbf{\Sigma}^{-1}_{b}(\mathbf{x}-\mathbf{\mu}_b)}{\sqrt{\mathbf{s}^T\mathbf{\Sigma}^{-1}_{b}\mathbf{s}}\sqrt{(\mathbf{x}-\mathbf{\mu}_b)^T\mathbf{\Sigma}^{-1}_{b}(\mathbf{x}-\mathbf{\mu}_b)}}.
    \label{eqn:ACE}
\end{equation}
We expand upon PaDiM by introducing ACE to the model framework. At inference, the model measures the ACE similarity between a sample and the target signature matrix in a whitened feature space given a mini-batch of images $B$\cite{Peeples_2022} as shown in Equation \ref{eqn:LACE}: 
\begin{equation}
\doublehat{D}_{ACE}=\doublehat{\mathbf{s}_c}^T{\doublehat{\mathbf{x}_n}}
    \label{eqn:LACE}
\end{equation}
where $\doublehat{\mathbf{s}_c}= \frac{\hat{\mathbf{s}}}{\lVert \hat{\mathbf{s}} \rVert}$, $\doublehat{\mathbf{x}}= \frac{\hat{\mathbf{x}}}{\lVert \hat{\mathbf{x}} \rVert}$, $\hat{\mathbf{s}}=\mathbf{D}^{-\frac{1}{2}}\mathbf{U}^{T}\mathbf{s}$, and $\hat{\mathbf{x}}=\mathbf{D}^{-\frac{1}{2}}\mathbf{U}^{T}(\mathbf{x}-\mathbf{\mu}_{b})$. Here $\mathbf{U}$ and $\mathbf{D}$ are the eigenvectors and eigenvalues of the inverse background covariance matrix, $\mathbf{\Sigma}^{-1}_{b}$, respectively.

   \begin{figure} [ht]
   \begin{center}
   \begin{tabular}{c} 
   \includegraphics[height=12cm]{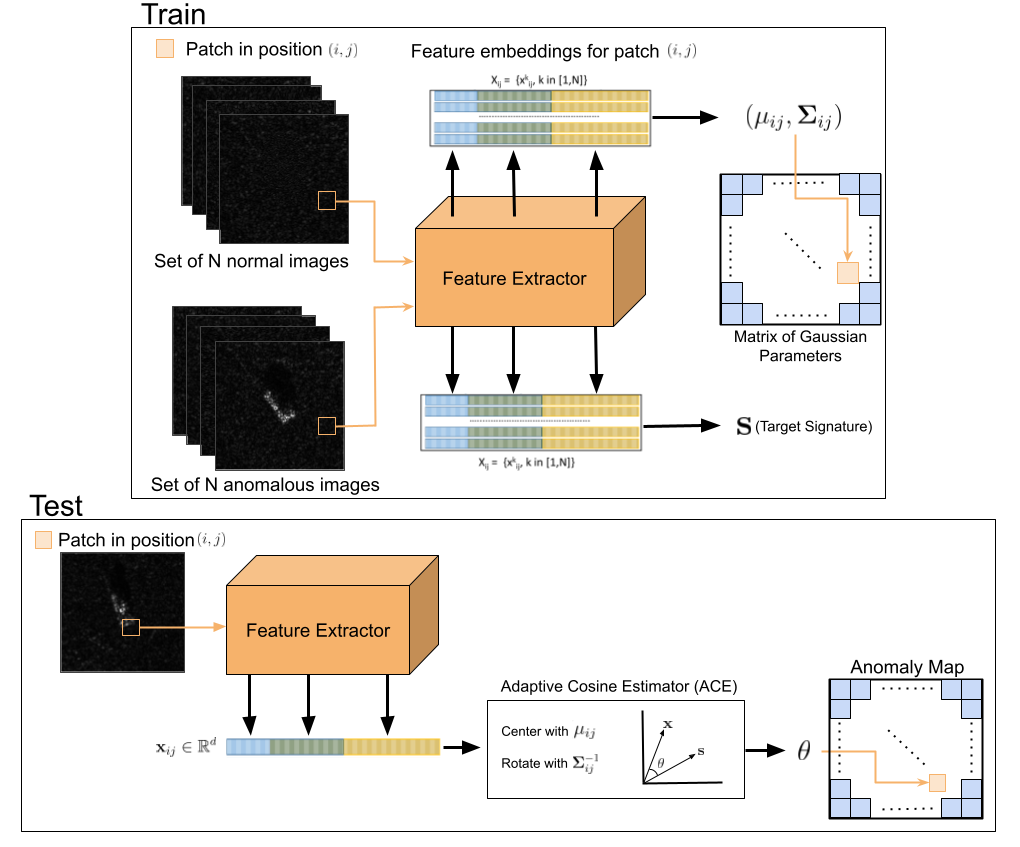}
   \end{tabular}
   \end{center}
   \caption[example] 
   { \label{fig:model-diagram} 
Normal and anomalous images are passed into the PaDiM framework at training. Normal image patches are used to create a matrix of Gaussian parameters while anomalous image patch embeddings are aggregated to create the class signature. At inference, a sample image is passed into the PaDiM framework and ACE is used to compute the similarity between a sample image patch and the class signature to create an anomaly map for each patch of the sample image. This figure is adapted from PaDiM \cite{defard2020padimpatchdistributionmodeling} for comparison.}
   \end{figure} 

Since anomaly detection is traditionally a binary classification problem, the background class distribution is generated from the normal image patch embeddings in the training step of the PaDiM framework. The target signature is generated by taking the mean of a set of anomalous image embeddings. The image patch embedding vectors are extracted using the same pre-trained CNN backbone as in the training step and randomly reduced to a size of $d$ features. Lastly, to create a more generalizable model, we further expand upon ACE and experiment with different covariance matrix types including full, diagonal, and isotropic matrices.

%% file: sections/Results_Discussion.tex
\section{Experimental Results and Discussion}
\label{sec:results_discussion}  

We conducted experiments on three different SAR datasets: moving and stationary target recognition (MSTAR) \cite{keydel1996mstar}, SAR Ship Detection Dataset (SSDD)\cite{rs13183690}, and High-Resolution SAR Images Dataset
(HRSID)\cite{9127939}. MSTAR is a widely used SAR dataset containing military ground vehicle targets captured at different aspect angles and configurations. For our experiments, we used the standard operating conditions set of MSTAR images. The normal images for MSTAR are created by removing the targets from the images using masks generated using k-Means clustering and replacing the anomalous regions with values randomly generated from the background distribution\cite{chauvin2025benchmarking}. SSDD is a ship detection dataset consisting of various onshore and offshore SAR images. The normal images for SSDD are generated in the same fashion as the normal MSTAR images. HRSID is a high-resolution SAR image dataset for ship detection, containing multiple target sizes and complex backgrounds. The HRSID dataset provides SAR images with pure backgrounds (no anomalies present) which are used as the normal images. For all datasets, we create train, validation, and test splits. 

We follow Anomalib's default split ratio, where 80\% of the normal images are used for training, while the remaining 20\% are allocated to testing and validation. The validation and test sets are evenly divided, each containing 50\% of the remaining normal images and 50\% of the anomalous images. A pre-trained ResNet-18 model is used as the feature extraction backbone, and the extracted feature embeddings are randomly reduced to a size of 100. We evaluate our results with the Area Under the Receiver Operating Characteristic curve (AUROC) which measures the ability of the model to distinguish between normal and anomalous regions. We report image-level AUROC for all three datasets and pixel-level AUROC for the HRSID and SSDD datasets. 

\begin{figure}[H]
    \centering
    \includegraphics[width=0.7\linewidth]{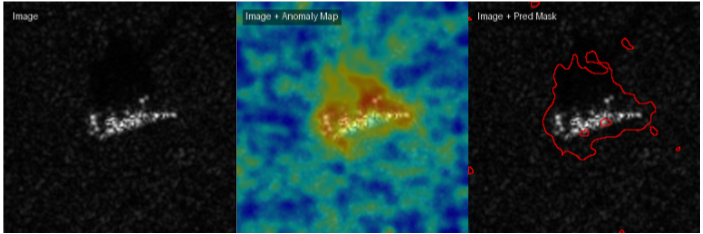}
    \caption{PaDiM-ACE result on the MSTAR dataset. The input image, anomaly heat map, and predicted mask are shown.}
    \label{fig:mstar_result}
\end{figure}

\begin{figure}[H]
    \centering
    \includegraphics[width=0.7\linewidth]{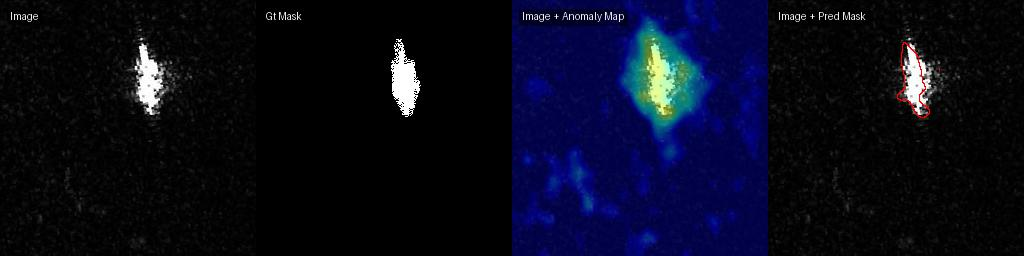}
    \caption{PaDiM-ACE result on the SSDD dataset. The input image, ground truth mask, anomaly heat map, and predicted mask are shown.}
    \label{fig:SSDD_result}
\end{figure}

We first compare our PaDiM-ACE model against other state of the art zero-shot anomaly detection models implemented in Anomalib including PaDiM\cite{defard2020padimpatchdistributionmodeling}, Deep Feature Modeling (DFM)\cite{ahuja2019probabilisticmodelingdeepfeatures}, and Window-based Contrastive Language–Image Pre-training (WinCLIP)\cite{jeong2023winclipzerofewshotanomalyclassification}. Table~\ref{tab:Comparison of models} presents a quantitative comparison of these methods across MSTAR, SSDD, and HRSID datasets. Our results show that PaDiM-ACE achieves competitive performance, particularly excelling in pixel-level anomaly detection for the SSDD dataset, as seen in Table~\ref{tab:Comparison of models} and Figure~\ref{fig:SSDD_result}. However, as illustrated in Figure~\ref{fig:mstar_result}, the model struggles with the MSTAR dataset, generating a higher number of false positives. This could be attributed to the imprecise mask annotations in MSTAR that include the shadow of the target in the masks, which may affect the quality of the generated normal images and, consequently, the background statistics used for anomaly detection.
\begin{table}[ht]
\caption{Comparison of PaDiM-ACE with other zero-shot image segmentation models in Anomalib. Results are displayed as a tuple (image AUROC, pixel AUROC). The average metrics are shown across three experimental runs and the standard deviation is provided. The best average metric is bolded.} 
\label{tab:Comparison of models}
\begin{center}  
\begin{tabular}{|c|c|c|c|}
\hline
\rule[-1ex]{0pt}{3.5ex}  Model & MSTAR & HRSID & SSDD \\
\hline
\rule[-1ex]{0pt}{3.5ex}  PaDiM-ACE & (84.99$\pm$2.56) & (71.79$\pm$7.42, 95.98$\pm$0.88) & (63.47$\pm$4.57, \textbf{94.25$\pm$4.56}) \\
\hline
\rule[-1ex]{0pt}{3.5ex}  PaDiM & (98.76$\pm$0.61) & (\textbf{87.24$\pm$3.22}, \textbf{97.04$\pm$0.72}) & (\textbf{68.97$\pm$4.41}, 93.65$\pm$1.02) \\
\hline
\rule[-1ex]{0pt}{3.5ex}  DFM & (99.09$\pm$0.69) & (63.21$\pm$3.12, 90.75$\pm$0.11) & (62.57$\pm$2.02, 90.98$\pm$0.91) \\
\hline
\rule[-1ex]{0pt}{3.5ex}  WinCLIP & (\textbf{99.62$\pm$0.22}) & (62.92$\pm$2.35, 85.28$\pm$1.24) & (51.22$\pm$1.20, 91.62$\pm$0.89) \\
\hline
\end{tabular}
\end{center}
\end{table}

\begin{figure}[H]
    \centering
    \includegraphics[width=0.7\linewidth]{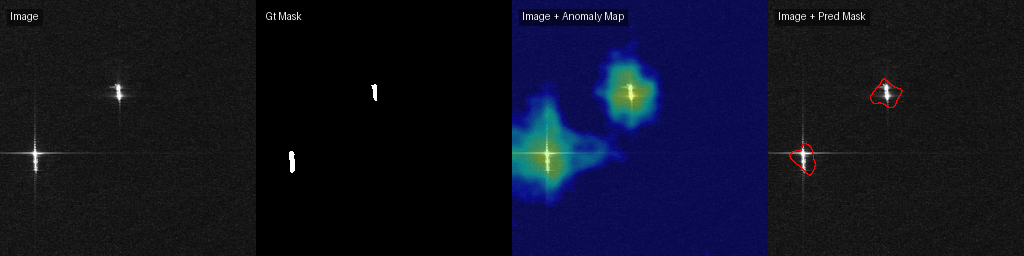}
    \caption{PaDiM-ACE result on the HRSID dataset. The input image, ground truth mask, anomaly heat map, and predicted mask are shown.}
    \label{fig:HRSID_result}
\end{figure}

Additionally, we investigate the impact of different feature extractors on our method by evaluating various backbones including: ResNet-18, ConvNeXT, Swin ViT, and MobileViT. The results are presented in Table~\ref{tab:Investigating different feature extractor backbones}. ResNet-18 provides the the highest scores for both image and pixel level metrics. However, the model does not perform well with vision transformer based architectures. This can most likely be attributed to the feature extraction process, which relies on embeddings extracted only from the first three layers of the model. This approach is better aligned with the architecture of CNNs rather than with vision transformers, which require different strategies for effective feature extraction. Furthermore, transformers typically require a large amount of data for effective representation learning. The limited training data available in our setting may have hindered their ability to generate adequate feature embeddings.
\begin{table}[ht]
\caption{Investigating different backbones in the PaDiM-ACE model using the HRSID dataset. The average metrics are shown across three experimental runs and the standard deviation is provided. The best average metric is bolded.} 
\label{tab:Investigating different feature extractor backbones}
\begin{center}       
\begin{tabular}{|c|c|c|c|c|} 
\hline
\rule[-1ex]{0pt}{3.5ex}   & Resnet18 & ConvNeXT & Swin ViT & MobileViT \\
\hline
\rule[-1ex]{0pt}{3.5ex}  Image AUROC & \textbf{68.74$\pm$2.92} & 61.62$\pm$6.43 & 53.88$\pm$8.97 & 55.36$\pm$5.30 \\
\hline
\rule[-1ex]{0pt}{3.5ex}  Pixel AUROC & \textbf{95.57$\pm$1.64} & 48.57$\pm$9.30 &  50.06$\pm$1.62& 15.52$\pm$13.66\\
\hline
\end{tabular}
\end{center}
\end{table}

We further examine the influence of different covariance matrix types used for background statistics in the ACE loss function. Table~\ref{tab:Investigating different cov mat} presents results for full, diagonal, and isotropic covariance matrices. The isotropic matrix is generated by computing the average variance of the diagonal background covariance matrix. The isotropic matrix achieves promising results suggesting that generalizing the model parameters enhances the performance while maintaining strong pixel-level AUROC results. These findings highlight the potential for extending PaDiM-ACE with a learnable adaptive cosine estimator (LACE)\cite{Peeples_2022} where rather than learning a full covariance matrix, a more compact representation could improve performance while reducing the number of parameters.
\begin{table}[ht]
\caption{Investigating different covariance matrix types for the background covariance matrix in PaDiM-ACE using the HRSID dataset. The average metrics are shown across three experimental runs and the standard deviation is provided. The best average metric is bolded.} 
\label{tab:Investigating different cov mat}
\begin{center}       
\begin{tabular}{|c|c|c|c|} 
\hline
\rule[-1ex]{0pt}{3.5ex}   & Full & Diagonal & Isotropic \\
\hline
\rule[-1ex]{0pt}{3.5ex}  Image AUROC & 73.79$\pm$4.91 & 47.52$\pm$5.00 & \textbf{81.29$\pm$2.23} \\
\hline
\rule[-1ex]{0pt}{3.5ex}  Pixel AUROC & \textbf{96.37$\pm$0.22} & 93.48$\pm$1.06 & 95.89$\pm$0.82 \\
\hline
\end{tabular}
\end{center}
\end{table}

Finally, we investigate the impact of various aggregation operations for the isotropic covariance matrix within the ACE loss function. Table~\ref{tab:Investigating different aggregation operations} presents a comparison of several aggregation methods, including computing the mean of the diagonal (average variance) covariance matrix, the mean of the full covariance matrix, the determinant of the covariance matrix, and the trace of the covariance matrix. Our results reveal that the determinant-based aggregation yields the highest image-level AUROC, while the mean of the full covariance matrix produces the best pixel-level AUROC. These differences in performance can likely be attributed to the ability of the determinant and full covariance aggregation to capture more complex feature relationships. In contrast, simpler methods, such as the mean of the diagonal or trace, assume more independent feature behaviors, which may not adequately model the dependencies among features.
\begin{table}[ht]
\caption{Investigating different aggregation operations for the isotropic covariance matrix in the PaDiM-ACE using the HRSID dataset. The average metrics are shown across three experimental runs and the standard deviation is provided. The best average metric is bolded.} 
\label{tab:Investigating different aggregation operations}
\begin{center}       
\begin{tabular}{|c|c|c|c|c|} 
\hline
\rule[-1ex]{0pt}{3.5ex}   & Mean of Diagonal & Mean of Full & Determinant & Trace  \\
\hline
\rule[-1ex]{0pt}{3.5ex}  Image AUROC & 79.64$\pm$4.52 & 77.75$\pm$6.35 & \textbf{81.68$\pm$8.34} & 65.17$\pm$5.40 \\
\hline
\rule[-1ex]{0pt}{3.5ex}  Pixel AUROC & 96.18$\pm$0.96 & \textbf{96.45$\pm$0.39} & 96.10$\pm$1.06 & 95.29$\pm$2.58 \\
\hline
\end{tabular}
\end{center}
\end{table}

%% file: sections/Conclusion.tex
\section{Conclusion}
\label{sec:conclusion}  

In this work, we introduced PaDiM-ACE, an extension of the PaDiM framework for anomaly detection and localization in SAR imagery. By introducing the ACE detection statistic, we achieved comparable performance to the original PaDiM framework, replacing the unbounded Mahalanobis distance with a bounded cosine similarity-based metric. This modification provides more robust anomaly scores, enhancing the model's ability to detect and localize anomalies within SAR images. Our experiments across multiple SAR datasets demonstrated that PaDiM-ACE provides competitive anomaly detection performance through a bounded cosine similarity metric. Additionally, we explored the impact of different feature extraction backbones and covariance matrix assumptions, highlighting the potential for further optimization. Future work will focus on refining the model’s robustness to SAR by introducing a framework that can learn target signature parameters and extend to a multiclass setting by leveraging approaches that build on the ACE metric \cite{Peeples_2022}. We could also adapt this approach to leverage physics-informed machine learning to directly tie the target and background statistics to the characteristics of SAR data.

%% file: main.bbl
\begin{thebibliography}{10}

\bibitem{10547107}
García, L.~P., Furano, G., Ghiglione, M., Zancan, V., Imbembo, E., Ilioudis, C., Clemente, C., and Trucco, P., ``Advancements in onboard processing of synthetic aperture radar (sar) data: Enhancing efficiency and real-time capabilities,'' {\em IEEE Journal of Selected Topics in Applied Earth Observations and Remote Sensing}~{\bf 17},  16625--16645 (2024).

\bibitem{liu2022high}
Liu, Y., Sun, G.-C., Guo, L., Xing, M., Yu, H., Fang, R., and Wang, S., ``High-resolution real-time imaging processing for spaceborne spotlight sar with curved orbit via subaperture coherent superposition in image domain,'' {\em IEEE Journal of Selected Topics in Applied Earth Observations and Remote Sensing}~{\bf 15},  1992--2003 (2022).

\bibitem{9852257}
Liu, G., Liu, B., Zheng, G., and Li, X., ``Environment monitoring of shanghai nanhui intertidal zone with dual-polarimetric sar data based on deep learning,'' {\em IEEE Transactions on Geoscience and Remote Sensing}~{\bf 60},  1--18 (2022).

\bibitem{10282026}
An, S. and Kim, D.-J., ``Sar simulation of phase and amplitude images enhancing military target identification,'' in [{\em IGARSS 2023 - 2023 IEEE International Geoscience and Remote Sensing Symposium}{\nolinebreak\hspace{0.1em}]},   7965--7968 (2023).

\bibitem{ALALI2022103295}
AlAli, Z.~T. and Alabady, S.~A., ``A survey of disaster management and sar operations using sensors and supporting techniques,'' {\em International Journal of Disaster Risk Reduction}~{\bf 82},  103295 (2022).

\bibitem{10283391}
Muzeau, M., Ren, C., Angelliaume, S., Datcu, M., and Ovarlez, J.-P., ``Self-supervised sar anomaly detection guided with rx detector,'' in [{\em IGARSS 2023 - 2023 IEEE International Geoscience and Remote Sensing Symposium}{\nolinebreak\hspace{0.1em}]},   1918--1921 (2023).

\bibitem{li2024sardet100kopensourcebenchmarktoolkit}
Li, Y., Li, X., Li, W., Hou, Q., Liu, L., Cheng, M.-M., and Yang, J., ``Sardet-100k: Towards open-source benchmark and toolkit for large-scale sar object detection,'' (2024).

\bibitem{defard2020padimpatchdistributionmodeling}
Defard, T., Setkov, A., Loesch, A., and Audigier, R., ``Padim: a patch distribution modeling framework for anomaly detection and localization,'' (2020).

\bibitem{Li_Deng_Li_Jiang_2019}
Li, X., Deng, S., Li, L., and Jiang, Y., ``Outlier detection based on robust mahalanobis distance and its application,'' {\em Open Journal of Statistics}~{\bf 09}(01),  15–26 (2019).

\bibitem{10730574}
Kesharwani, A. and Shukla, P., ``A review of anomaly detection using machine learning techniques,'' in [{\em 2024 1st International Conference on Advanced Computing and Emerging Technologies (ACET)}{\nolinebreak\hspace{0.1em}]},   1--6 (2024).

\bibitem{akcay2022anomalib}
Akcay, S., Ameln, D., Vaidya, A., Lakshmanan, B., Ahuja, N., and Genc, U., ``Anomalib: A deep learning library for anomaly detection,'' in [{\em 2022 IEEE International Conference on Image Processing (ICIP)}{\nolinebreak\hspace{0.1em}]},   1706--1710, IEEE (2022).

\bibitem{1990ITASS..38.1760R}
{Reed}, I.~S. and {Yu}, X., ``{Adaptive multiple-band CFAR detection of an optical pattern with unknown spectral distribution},'' {\em IEEE Transactions on Acoustics Speech and Signal Processing}~{\bf 38},  1760--1770 (Oct. 1990).

\bibitem{9987646}
Muzeau, M., Ren, C., Angelliaume, S., Datcu, M., and Ovarlez, J.-P., ``Self-supervised learning based anomaly detection in synthetic aperture radar imaging,'' {\em IEEE Open Journal of Signal Processing}~{\bf 3},  440--449 (2022).

\bibitem{Peeples_2022}
Peeples, J., McCurley, C.~H., Walker, S., Stewart, D., and Zare, A., ``Learnable adaptive cosine estimator (lace) for image classification,'' in [{\em 2022 IEEE/CVF Winter Conference on Applications of Computer Vision (WACV)}{\nolinebreak\hspace{0.1em}]},   3757–3767, IEEE (Jan. 2022).

\bibitem{keydel1996mstar}
Keydel, E.~R., Lee, S.~W., and Moore, J.~T., ``Mstar extended operating conditions: A tutorial,'' {\em Algorithms for Synthetic Aperture Radar Imagery III}~{\bf 2757},  228--242 (1996).

\bibitem{rs13183690}
Zhang, T., Zhang, X., Li, J., Xu, X., Wang, B., Zhan, X., Xu, Y., Ke, X., Zeng, T., Su, H., Ahmad, I., Pan, D., Liu, C., Zhou, Y., Shi, J., and Wei, S., ``Sar ship detection dataset (ssdd): Official release and comprehensive data analysis,'' {\em Remote Sensing}~{\bf 13}(18) (2021).

\bibitem{9127939}
Wei, S., Zeng, X., Qu, Q., Wang, M., Su, H., and Shi, J., ``Hrsid: A high-resolution sar images dataset for ship detection and instance segmentation,'' {\em IEEE Access}~{\bf 8},  120234--120254 (2020).

\bibitem{chauvin2025benchmarking}
Chauvin, L., Gupta, S., Ibarra, A., and Peeples, J., ``Benchmarking suite for synthetic aperture radar imagery anomaly detection (sariad) algorithms,'' in [{\em Algorithms for Synthetic Aperture Radar Imagery XXXII}{\nolinebreak\hspace{0.1em}]},   {\bf 13456}, SPIE (2025).

\bibitem{ahuja2019probabilisticmodelingdeepfeatures}
Ahuja, N.~A., Ndiour, I., Kalyanpur, T., and Tickoo, O., ``Probabilistic modeling of deep features for out-of-distribution and adversarial detection,'' (2019).

\bibitem{jeong2023winclipzerofewshotanomalyclassification}
Jeong, J., Zou, Y., Kim, T., Zhang, D., Ravichandran, A., and Dabeer, O., ``Winclip: Zero-/few-shot anomaly classification and segmentation,'' (2023).

\end{thebibliography}
